\documentclass[oribibl]{llncs}
\usepackage{llncsdoc}
\usepackage{subfig}
\usepackage{setspace,tabularx,booktabs,ragged2e,ifthen,amsmath, hyperref, amssymb, graphicx,amsfonts} 
\usepackage{algorithm}
\usepackage{algorithmic}
\newcommand*\Bell{\ensuremath{\boldsymbol\ell}}
\begin{document}
\title{Fast Spectral Clustering \\ Using Autoencoders and Landmarks}
\author{Ershad Banijamali\inst{1} \and Ali Ghodsi\inst{2}}

\institute{School of Computer Science, University of Waterloo, Canada
\and Department of Statistics and Actuarial Science, University of Waterloo, Canada
}
\maketitle

\begin{abstract} 
In this paper, we introduce an algorithm for performing spectral clustering efficiently. Spectral clustering is a powerful clustering algorithm that suffers from high computational complexity, due to eigen decomposition. In this work, we first build the adjacency matrix of the corresponding graph of the dataset. To build this matrix, we only consider a limited number of points, called landmarks, and compute the similarity of all data points with the landmarks. Then, we present a definition of the Laplacian matrix of the graph that enable us to perform eigen decomposition efficiently, using a deep autoencoder. The overall complexity of the algorithm for eigen decomposition is $O(np)$, where $n$ is the number of data points and $p$  is the number of landmarks. At last, we evaluate the performance of the algorithm in different experiments. 

\end{abstract} 
\section{Introduction}
Clustering is a long-standing problem in statistical machine learning and data mining. Many different approaches have been introduced in the past decades to tackle this problem. Spectral clustering is one of the most powerful tools for clustering. The idea of spectral clustering originally comes from the min-cut problem in graph theory. In fact, if we represent a dataset by a graph where the vertices are the data points and edges are the similarity between them, then the final clusters of spectral clustering are the cliques of the graph that are formed by cutting minimum number of the edges (or minimum total weight in weighted graphs). 

Spectral clustering is capable of producing very good clustering results for small datasets. It has application in different fields of  data analysis and machine learning \cite{dhillon2001co,ng2001spectral,paccanaro2006spectral,white2005spectral}. However, the main drawback of this algorithm comes from the eigen decomposition step, which is computationally expensive ($O(n^3)$, $n$ being the number of data points). To solve this problem, many algorithms have been designed. These algorithms are mainly based on sampling from the data (or the affinity matrix), solving the problem for the samples, and reconstruction of the solution for the whole dataset based on the solution for the samples. In \cite{williams2000using,fowlkes2004spectral,li2011time}, Nystrom method has been used to sample columns from affinity matrix and the the full matrix is approximated using correlation between the selected columns and the remaining columns. In \cite{choromanska2013fast}, a  performance guarantee for these approaches has been derived and a set of conditions have been discussed under which this approximation performs comparable to the exact solution. \cite{gittens2013approximate} suggests an iterative process for approximating the eigenvector of the Laplacian matrix. In \cite{yan2009fast}, $k$-means or random projection is used to find centroids of the partitions of the data. Then, they perform spectral clustering on the centroid, and finally, assign each data point to the cluster of their corresponding centroids.  

The most relevant work to ours, however, is by Chen et al. \cite{chen2011large}. The authors proposed a method for accelerating the spectral clustering based on choosing $p$ landmarks in the dataset and computing the similarities between all points and these landmarks to form a $p \times n$ matrix. Then the eigenvectors of the full matrix is approximated by the eigendecomposition of this $p \times n$ matrix. The overall complexity of this method is $O(np^2+p^3)$. The results of the method are close to the actual spectral clustering on $n$ data points, but with much less computation time. 

Multi-layer structures have been used for spectral clustering in some recent works \cite{shao2015deep,tian2014learning}. Training deep architectures is done much faster than eigendecomposition, since it can be easily parallelized over multiple cores. However, in the mentioned works, the size of input layer of the network is equal to the number of data points, i.e. $n$, and consequently the whole network is drastically enlarged as $n$ grows. Therefore, it will be infeasible to use these structures for large datasets. In this paper, we combine the idea of landmark selection and deep structures to achieve a fast and yet accurate clustering. The overall computational complexity of the algorithm, given the parallelization of the network training, is $O(np)$.
\vspace{-.2cm}
\section{Background:}
\subsection{Spectral Clustering}
Mathematically speaking, suppose we have a dataset $\mathbf{X}$ with $n$  data points, $\{\mathbf{x}_1,\mathbf{x}_2,...,\mathbf{x}_n \}$. We want to partition this set to $k$ clusters. To do spectral clustering, we first form the corresponding graph of the dataset, where the vertices are the points, and then obtain the adjacency matrix of the graph, denoted by $W$. Each entry of $W$ shows the similarity between each pair of points. So $W$ is a symmetric matrix. The degree matrix of the graph, denoted by $D$, is a diagonal matrix and its nonzero elements are summation over all the elements of rows (or columns) of $W$, $d_{ii} = \sum_j w_{ij}  $.

Based on $D$ and $W$, the Laplacian matrix of the graph, denoted by $L$, is obtained. There are different ways for defining the Laplacian matrix. The unnormalized Laplacian matrix is defined as: $L = D - W$. To do spectral clustering, we can get the final clusters of the dataset, by running $k$-means on the $k$ eigenvectors of $L$, corresponding to the $k$ smallest eigenvalues (the smallest eigenvalue of $L$ is $0$ and its corresponding eigenvector, which is constant, is discarded). The are some normalized versions of $L$, which usually yield a better clustering results. One of them is defined as $L_{norm} = D^{-1/2} L D^{-1/2}$ \cite{chung1997spectral}. We can also use the $k$ eigenvectors corresponding to the $k$ smallest eigenvalues of $L_{norm}$ (or equivalently $k$ largest eigenvalues of $L_{norm} = D^{-1/2} W D^{-1/2}$, according to \cite{ng2002spectral}).

\subsection{Autoencoders and eigendecomposition}
The relation between autoencoder and eigendecomposition was first revealed in Hinton and Salakhutdinov's paper \cite{hinton2006reducing}. Consider the Principal Component Analysis (PCA) problem. We would like to find a low-dimensional representation of data, denoted by $Z$, which preserves the variation of the original data, denoted by $X$, as much as possible. Suppose $U$ is the linear transformation for PCA problem, which projects the data points in a low-dimensional space, $Z = U^{\top}X$. To keep the maximum variation of the data, it is known that, the basis of the low-dimensional space (or columns of matrix $U$) are the eigenvectors of the covariance matrix, $\text{Cov}_X=XX^{\top}$, corresponding to the largest eigenvalues. PCA can also interpreted as finding a linear transformation that minimizes the reconstruction loss, i.e. 

\begin{equation}
\min_U \parallel X - UZ \parallel^2
\end{equation}

The above objective function is exactly used in conventional autoencoders. In fact, a single layer autoencoder with no nonlinearity spans the same low-dimensional space as its latent space. However, deep autoencoders are capable of finding better low-dimensional representations than PCA.  

Training an autoencoder using backpropagation is much faster than solving the eigen decomposition problem. In \cite{tian2014learning}, the authors used autoencoders for  graph clustering. Instead of using an actual data point as the input of the autoencoder, they use vector of similarity of that point with other points. Their results show benefit of the model compared to some rival model, in terms of Normalized Mutual Information (NMI) criterion. However, since the length of the similarity vector is equal to the number of data points, $n$, extending the idea of this work for large datasets, with hundred of thousands or even million data points, is not feasible. 

Inspired by these works, we introduce a simple, but fast and efficient algorithm for spectral clustering using autoencoders. In the next section we describe the model. 
\section{Model Description}
As described in the previous section, spectral clustering can be done by decomposing the eigenvalues and eigenvectors of $L_{norm} = D^{-1/2} W D^{-1/2}$. In our work, we do this decomposition using an autoencoder. Instead of original feature vectors, we represent each data point by its similarity to other data points. However, instead of calculating the similarity of a given data point with all other data points and forming a vector of length $n$, we only consider some landmarks and compute the similarity of the points with these landmarks. Lets denote the $p$ selected landmarks ($p \ll n$) by $\{\Bell_1,\Bell_2,...,\Bell_n\}$. Then we compute the similarity of all data points with those landmarks and form a $p \times n$ matrix $W$. In this work, we used Gaussian kernel as the similarity measure between the points, i.e.:

\begin{equation}
w_{ij} = \exp(-\frac{\parallel \Bell_i - \mathbf{x}_j \parallel^2}{\sigma})
\end{equation} 
where $\sigma$ is the parameter of the model. To make the model more robust, we always set  $\sigma$ to be the median of the distance between data points and landmarks.

\begin{equation}
\sigma = \text{median} \{\parallel \Bell_i - \mathbf{x}_j \parallel^2 \}_{i,j = 1}^{p,n}
\end{equation}
This way, we also guarantee that the value of similarities are well spread in $[0,1]$ interval. Each column of matrix $W$, denoted by $\mathbf{w}_i$'s, represents a data point in the original set $X$ based on its similarity to the landmarks. Constructing matrix $W$ takes $O(npd)$ ($d$ being the number of features), which is inevitable in all similar algorithms. However, our main contribution is in decreasing computational cost in decomposition step.

Next, we have to form the Laplacian matrix. We should notice that $L_{norm}=D^{-1/2}WD^{-1/2}$ is no longer a valid matrix, since $W$ is $p \times n$. Now, we are looking for a Laplacian matrix that can be written in the form of $L_{norm} = SS^{\top}$, so that we can use $S$ as the input to our autoencoder in order to eigen decomposition. To do so, we define another matrix $M = W^{\top}W$. Based on this definition, $M$ is also a similarity matrix over the data points. However, since $m_{ij}= \mathbf{w}_i^{\top}\mathbf{w}_j$, $M$ is a more local measure than $W$, which is a good property for spectral clustering.

The diagonal matrix $D$ can be obtained by summing over elements in columns of $M$. However, since our goal in this work is to minimize the computational cost of the algorithm, we would like to avoid computing $M$, directly, which has a computation cost of $O(n^2p)$. Instead, we compute $D$ another way. In fact, we know $d_{ii} = \sum_i m_{ij} = \sum_i \mathbf{w}_i^{\top}\mathbf{w}_j$. We can write this as follows:
\begin{equation}
\begin{array}{c}
d_{ii} = \sum \limits_{j=1}^n  m_{ij} = \sum \limits_{j=1}^n \mathbf{w}_i^{\top}\mathbf{w}_j=  \mathbf{w}_i^{\top} \sum \limits_{j=1}^n \mathbf{w}_j = \mathbf{w}_i^{\top}\mathbf{w}^s
\end{array}
\end{equation}
where $\mathbf{w}^s$ is a $p \times 1$ vector and its $k$th elements is a sum over elements in the $k$th row of $W$. Therefore, $D$ can be written as:
\begin{equation}
\label{eq: degree}
D = diag(W^{\top} \mathbf{w}^s)
\end{equation}
Note that calculation of $\mathbf{w}^s$ and $D$ this way has complexity of $O(np)+O(np) = O(np)$, which is a significant improvement compared to $O(n^2p)$. 

We can then obtain the Laplacian matrix:  
\begin{equation}
\L_{norm} = D^{-1/2} M D^{-1/2} =  D^{-1/2} W^{\top}W D^{-1/2}
\end{equation}

By putting $S = W D^{-1/2}$ as the input of our autoencoder, we can start training the network. The objective function for training the autoencoder is minimizing the error of reconstructing $S$. After training the network, we obtain the representation of all data points in the latent space and run $k$-means on the latent space. Again, instead of computing $W D^{-1/2}$, we can simply multiply each diagonal element of $D^{-1/2}$ by the corresponding column of $W$, i.e. $\mathbf{s}_i = d_{ii}^{-1/2} \mathbf{w}_i$. This operation also has computational complextiy $O(np)$.

\begin{figure}[!t]
    \centering
\includegraphics[trim = 0mm 0mm 0mm 0mm,width=9cm]{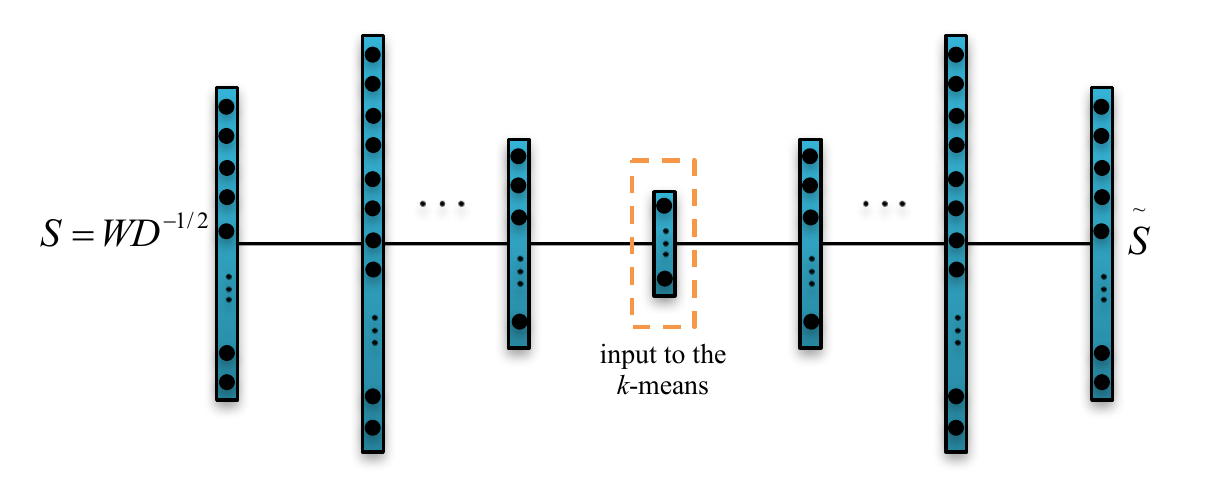} 
\caption{Input to the network is $WD^{-1/2}$ and $k$-means is performed in the latent space}  
\label{fig: autoencoder}
\vspace{-.5cm}
\end{figure}

Figure \ref{fig: autoencoder} shows the proposed model. The objective function for training the autoencoder, as described above, is to minimize the euclidean distance between the input and the output of the network, i.e. $S$ and $\widetilde{S}$. Training a network can be done very efficiently using backpropagation and mini-batch gradient descent, if the number of hidden units in each layer be in order of $p$, which is the case. Furthermore, in contrast to eigen decomposition problem, the training phase can be easily distributed over several machines (or cores).  These two facts together helps us to keep the computational complexity of decomposition step in $O(np)$. 

Algorithm \ref{alg:spectral clustering} describes the steps of the proposed method. Note that the $p$ landmarks can be obtained in different ways. They can be randomly sampled from the original dataset or be the centroids of $p$ clusters of the dataset by running $k$-means or be picked using column subset selection methods, e.g. \cite{farahat2015greedy}. 

\begin{algorithm}[h]
   \caption{Spectral Clustering using Autoencoders and Landmarks}
   \label{alg:spectral clustering}
\begin{algorithmic}
   \STATE \textbf{Input:} Dataset $\mathbf{X}$ with $n$ samples $\{\mathbf{x}_1, \mathbf{x}_2,...,\mathbf{x}_n \}$
   \STATE \textbf{Output:} $k$ clusters of the dataset
   \STATE 1: Select $p$ landmarks 
   \STATE 2: Compute the similarities between data points and landmarks and store \\ \hspace{.3cm} them in matrix $W$
   \STATE 3: Compute the degree matrix: $D = diag(W^{\top} \mathbf{w}^s)$
   \STATE 4: Compute  $S$, the input to the autoencoder: $\mathbf{s}_i = d_{ii}^{-1/2}\mathbf{w}_i$
   \STATE 5: Train an autoencoder using $S$ as its input
   \STATE 6: Run $k$-means on the latent space of the trained autoencoder
\end{algorithmic}
\end{algorithm}

\newpage

\section{Experiment Results}
\vspace{-.3cm}
In the following two subsections, we present the results of applying our clustering algorithm on different sets of data. In all of these experiments, the autoencoder has $5$ hidden layers between input and output layer. Only number of units in the layers changes for different datasets. The activation function for all hidden layer is ReLU, except the middle layer that has linear activation. The activation for output layer is sigmoid.
\vspace{-.3cm}
\subsection{Toy Datasets}
To demonstrate the performance of the proposed method, we first show the results for some small 2-dimensional datasets. Figure \ref{fig:toy} shows the performance of the algorithm on four different datasets. As we can see in this figure, the natural clusters of the data have been detected by a high accuracy. Number of landmarks for all of these experiments is set to $200$, and they are drawn randomly from the datasets. For all of these experiments, number of units in the hidden layers is: $64$, $32$, $2$, $32$, and $64$, respectively. 
\vspace{-.5cm}
\begin{figure}[!h]
\vspace{-.5cm}
    \centering
    \subfloat[]{{\includegraphics[trim = 0mm 60mm 0mm 60mm,width=3cm]{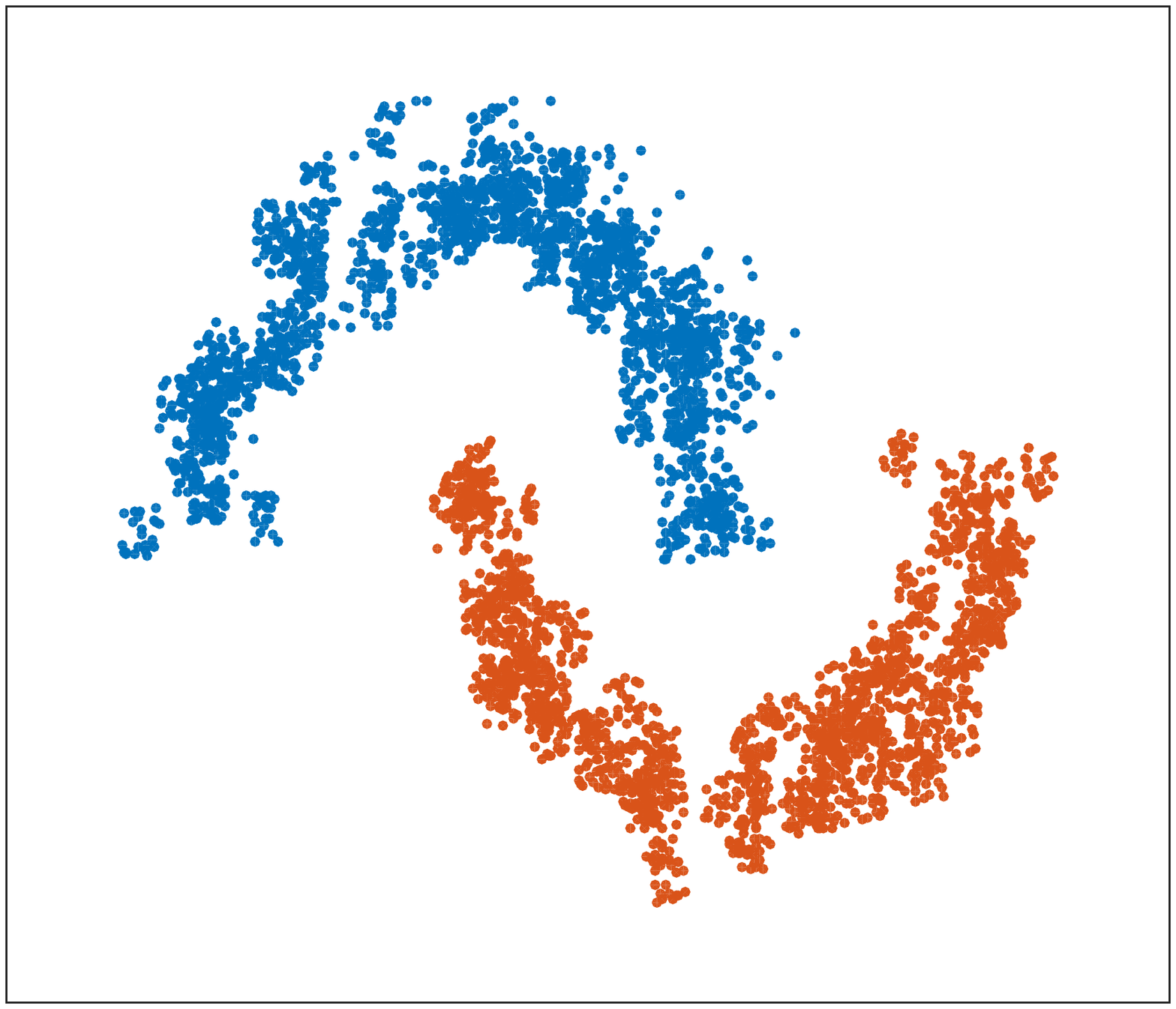} }}
	\subfloat[]{{\includegraphics[trim = 0mm 60mm 0mm 60mm,width=3cm]{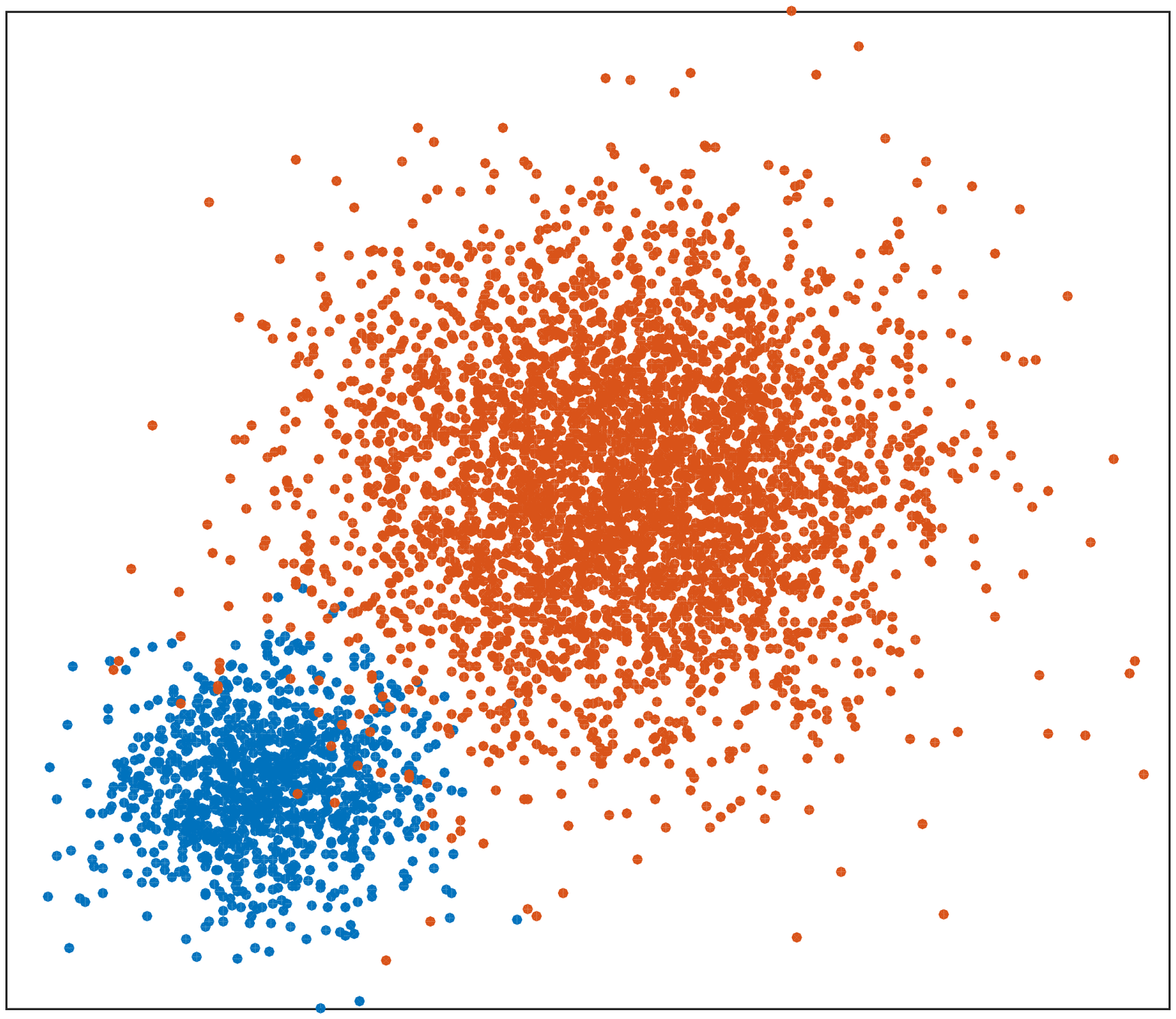} }} 	    
    \subfloat[]{{\includegraphics[trim = 0mm 60mm 0mm 60mm,width=3cm]{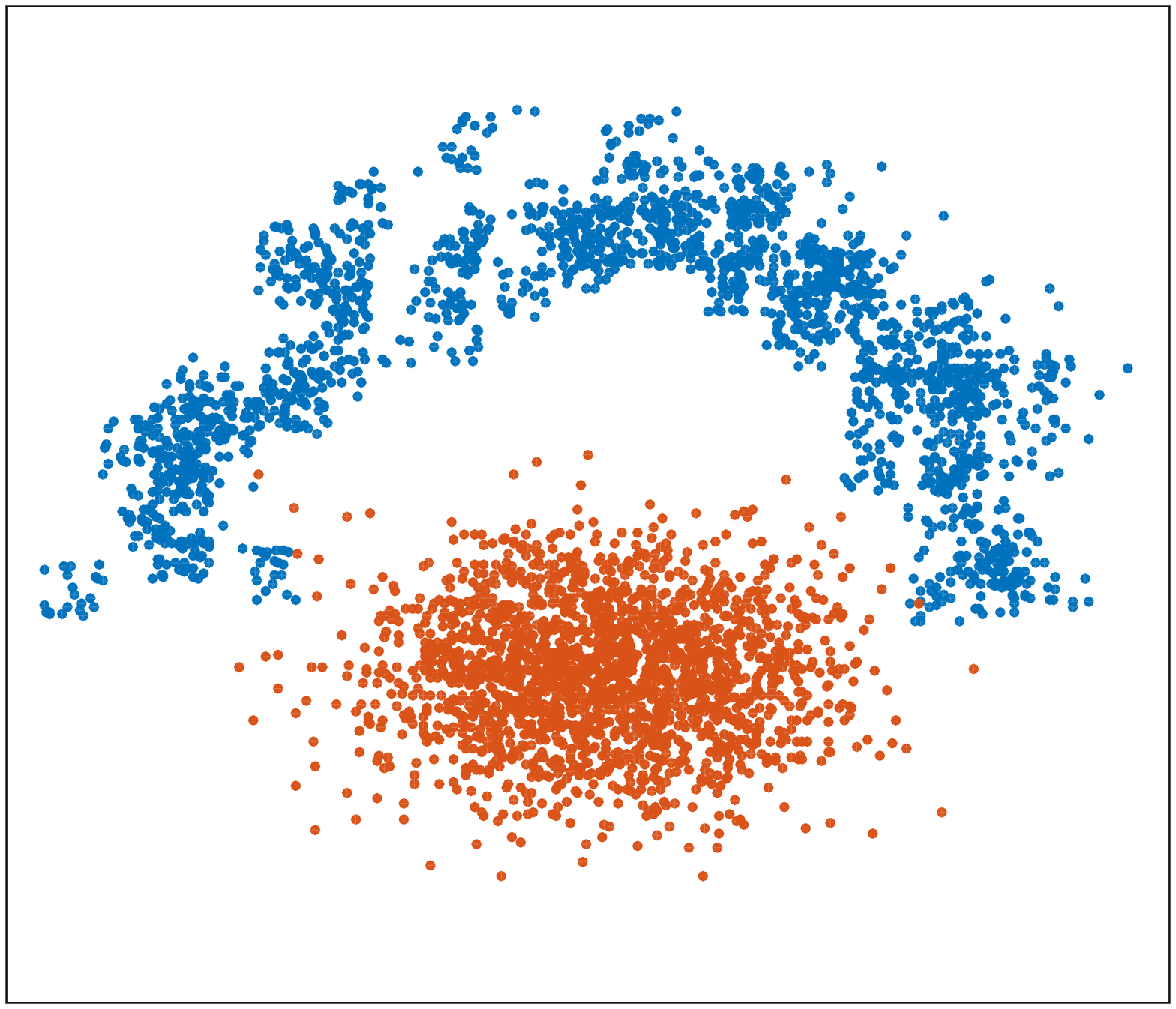} }}		
	\subfloat[]{{\includegraphics[trim = 0mm 60mm 0mm 60mm,width=3cm]{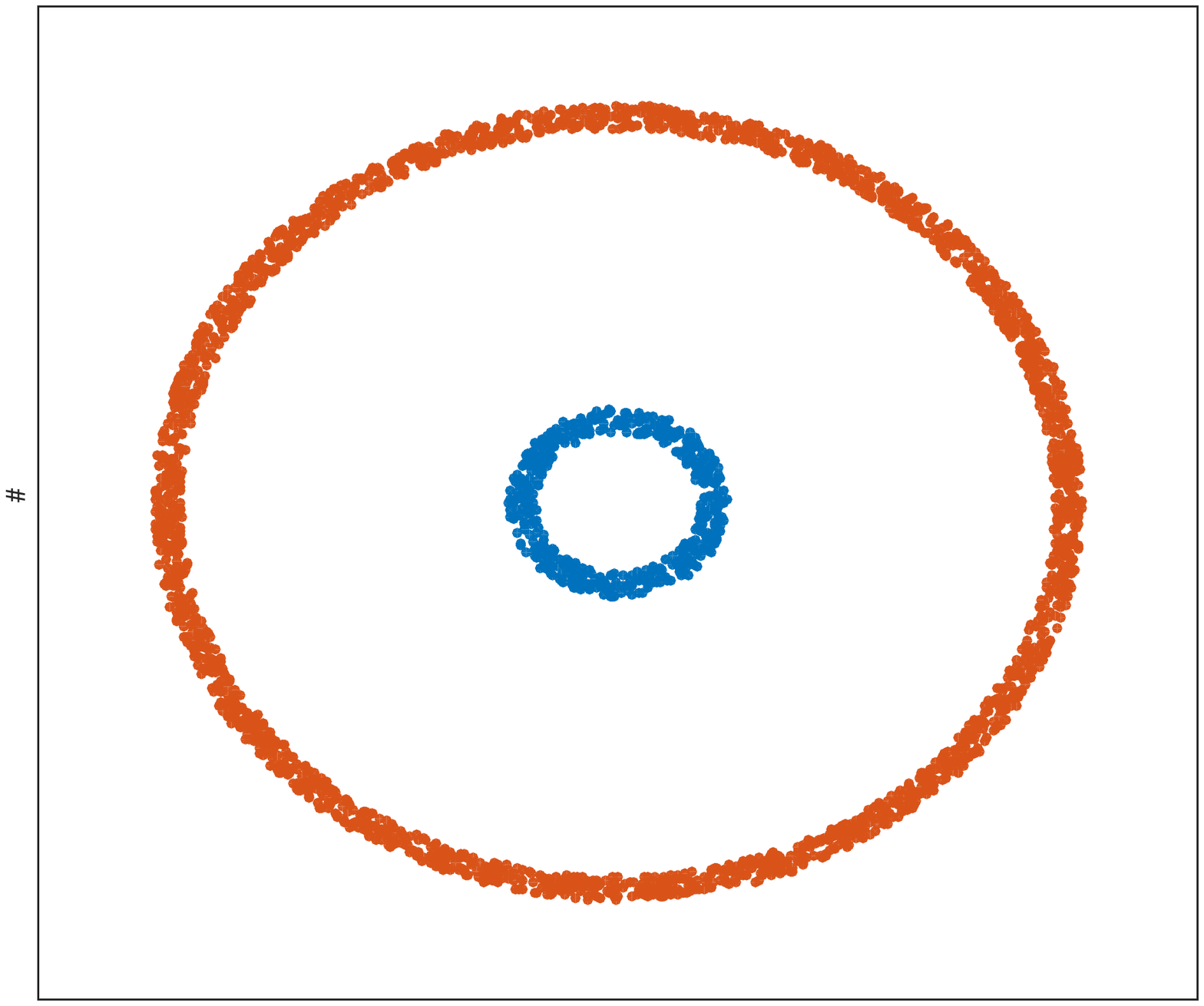} }}

        \vspace{-.2cm}
    \caption{Clustering using the proposed method: (a) Two-moon dataset, $n=4000$ (b) Two-circle dataset $n=4500$ (c) Moon-circle dataset $n=4000$ (d) Concentric rings $n=3000$} \vspace{-.6cm}
    \label{fig:toy}%
\end{figure}
\vspace{-.5cm}
\subsection{Real-World Datasets}
\vspace{-.3cm}
In this section we evaluate the proposed algorithm with more challenging and larger datasets. To measure the performance here, we use Clustering Purity (CP) criterion. CP is defined for a labeled dataset as a measure of matching between classes and clusters. If $\{C^1,C^2,...,C^L\}$ are $L$ classes of a dataset $\mathbf{X}$ of size $n$, then a clustering algorithm, $\mathcal{A}$, which divides $\mathbf{X}$ into $K$ clusters $\{X^1,X^2,...,X^K\}$ has $\textit{CP}(\mathcal{A},X)$ as:
\vspace{-.15cm}
\begin{equation}
\textit{CP}(\mathcal{A},X) = \frac{1}{n} \sum \limits_{j=1}^K \max \limits_i |C^i \cap X^j|. 
\end{equation}

Table \ref{tbl:datasets} contains a short description about each of the datasets. 

\begin{table}[!h]
\begin{center}
\begin{small}
\begin{tabular}{|l|c|c|l|}
\hline
Dataset & size ($n$)& $\#$ of classes & Description \\
\hline
MNIST 		& $60000$ & $10$ & $28 \times 28$ grayscale images of digits\\
Seismic 	& $98528$ & $3$  & Types of moving vehicle in a wireless sensor network\\
CIFAR-10 	& $50000$ & $10$ & $32 \times 32$ colored images of $10$ different objects\\
LetterRec 	& $20000$ & $26$ & Capital letters in English alphabet\\
\hline
\end{tabular}
\end{small}
\end{center}
\caption{Specification of the datasets}
\label{tbl:datasets}
\vspace{-1cm}
\end{table}

In table \ref{tbl:results}, we can compare the performance of our proposed algorithm, SCAL (Spectral Clustering with Autoencdoer and Landmarks), with some other clustering algorithms. SCAL has two variants: 1) SCAL-R where landmarks are selected randomly, 2) SCAL-K where landmarks are centroids of $k$-means. LSC-R and LSC-K are methods from \cite{chen2011large}, which also uses landmarks for spectral clustering. Based on this table, SCAL-K outperforms SCAL-R in almost all cases. As we increase the number of landmarks, the performance improves, and in some cases we get better result than original spectral clustering, which is an interesting observation. This may suggest that deep autoencoders are able to extract features that are more useful for clustering, compared to shallow structures. Another observation is that when we increase $p$, gap between SCAL-R and SCAL-K becomes smaller. This suggests that if we choose $p$ to be large enough (but still much smaller than $n$), even random selection of the landmark does not degrade the performance too much. 
\vspace{-.5cm}
\begin{table}[!h]
\begin{center}
{\tabcolsep=0pt\def\arraystretch{1}
\begin{tabularx}{300pt}{|l | *4{>{\Centering}X}|}
\hline
Algorithm &  MNIST &  Seismic & CIFAR-10 & LetterRec\tabularnewline \hline
Spectral Clustering 		& $71.54$ & $66.68$ & $\textbf{60.13}$ & $33.19$\tabularnewline
$k$-means  					& $57.31$ & $62.82$ & $40.12$ & $30.01$ \tabularnewline
LSC-R ($p=500$)				& $62.94$ & $66.19$ & $47.16$ & $29.44$\tabularnewline
LSC-K 	($p=500$)			& $68.10$ & $67.71$ & $50.40$ & $31.59$\tabularnewline
SCAL-R ($p=500$)			& $64.13$ & $64.41$ & $49.41$ & $29.52$\tabularnewline
SCAL-K ($p=500$)			& $69.14$ & $68.43$ & $54.64$ & $32.88$\tabularnewline
SCAL-R ($p=1000$) 			& $70.61$ & $67.55$ & $56.19$ & $33.94$\tabularnewline
SCAL-K ($p=1000$)			& $\textbf{72.98}$ & $\textbf{68.61}$ & $58.02$ & $\textbf{34.70}$\tabularnewline
\hline
\end{tabularx}}
\end{center}
\vspace{-.1cm}
\caption{Performance of different clustering algorithms in terms of CP. $p$ shows the number of landmarks. For all of these result we used $10$ epochs of data.}
\label{tbl:results}
\vspace{-1cm}
\end{table}

Figures below show the performance and runtime of the algorithm versus LSC-R and LSC-K methods, as a function of number of landmarks.
\vspace{-.8cm}
\begin{figure}[!h]
    \centering
    \subfloat[]{{\includegraphics[trim = 0mm 50mm 0mm 60mm,width=5.8cm]{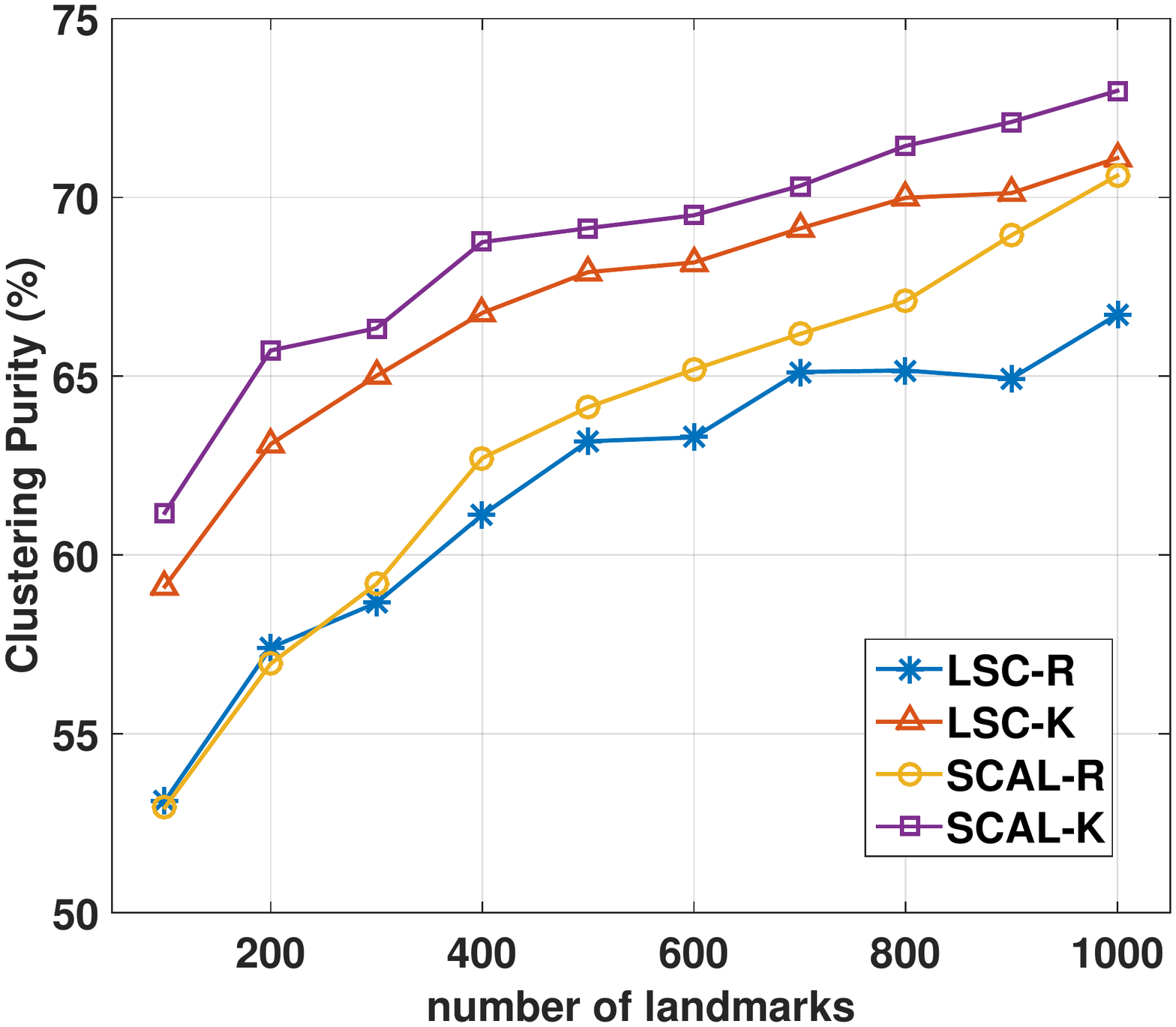} }}
	\subfloat[]{{\includegraphics[trim = 0mm 50mm 0mm 60mm,width=5.8cm]{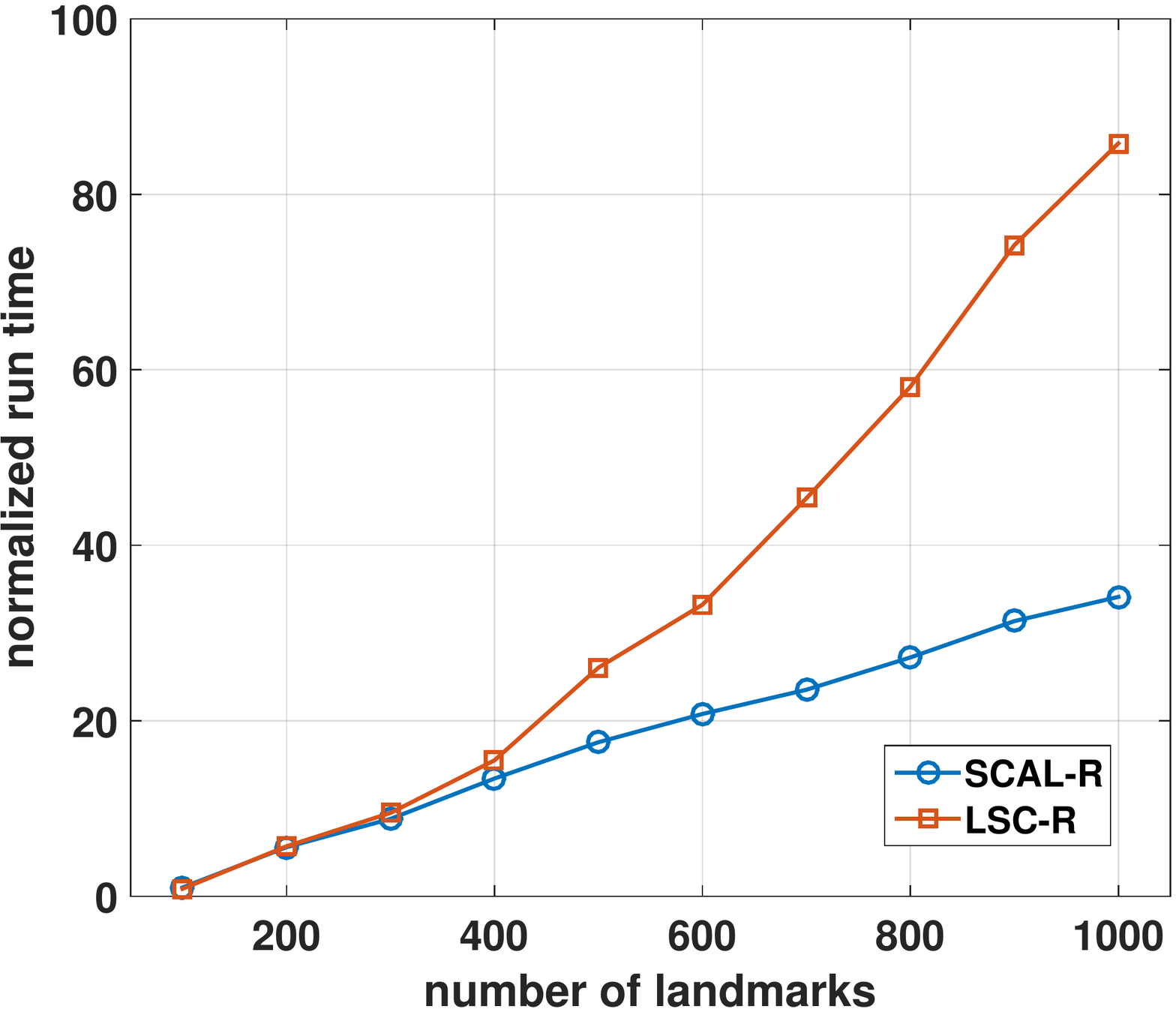} }} 	
	\vspace{-.3cm}   
    \caption{Performance of different methods versus number of landmarks (a) clustering purity (b) normalized run time; SCAL-K and LSC-K have similar behavior, since they just have an additional overhead} \vspace{-.6cm}
    \label{fig:toy}%
\end{figure}

\newpage
\section{Conclusion}
\vspace{-.4cm}
We introduced a novel algorithm using landmarks and deep autoencoders, to perform spectral clustering efficiently. The complexity of the algorithm is $O(np)$, which is much faster than the original spectral clustering algorithm as well as some other approximation methods. Our experiment shows that, despite the gain in computation speed, there is no or limited loss in clustering performance. 
\vspace{-.5cm}
\bibliography{Spectral_Clustering}
\bibliographystyle{splncs}

\end{document}